\documentclass[10pt,journal,compsoc]{IEEEtran}

\ifCLASSOPTIONcompsoc
\usepackage{subcaption}
\usepackage{graphicx}
\usepackage{caption}
\usepackage{multirow}
\usepackage{adjustbox}
\usepackage{tabularx}
\usepackage{amsmath}
\usepackage{amssymb}
\usepackage{color}
\usepackage{algorithm}
\usepackage{algpseudocode}
\usepackage{scrextend}
\usepackage{todonotes}
\usepackage{lipsum}
\usepackage[nocompress]{cite}
\else
 \usepackage{cite}
\fi
\ifCLASSINFOpdf
\else
\fi

\nocite{*}

\def\bs{\boldsymbol}

\begin{document}
\title{A Nonlinear Dimensionality Reduction Framework Using Smooth Geodesics}

\author{Kelum~Gajamannage, Randy~Paffenroth, and Erik~M. Bollt
\IEEEcompsocitemizethanks{\IEEEcompsocthanksitem K. Gajamannage is with the Department
of Mathematical Sciences, Worcester Polytechnic Institute, Worcester, MA, 01609.\protect\\
E-mail: kdgajamannage@wpi.edu
\IEEEcompsocthanksitem R. Paffenroth is with the Department
of Mathematical Sciences, Department of Computer Science, and Data Science Program, Worcester Polytechnic Institute, Worcester, MA, 01609.\protect\\
E-mail:  rcpaffenroth@wpi.edu
\IEEEcompsocthanksitem E. M. Bollt is with with the Clarkson Center for Complex Systems Science, Clarkson University, Potsdam, NY, 13676.\protect\\
E-mail:  ebollt@clarkson.edu}
\thanks{Manuscript received MONTH DATE, YEAR; revised August MONTH DATE, YEAR.}}

\markboth{~Vol.~XX, No.~XX, MONTH~YEAR}%
{Shell \MakeLowercase{\textit{et al.}}: Bare Advanced Demo of IEEEtran.cls for IEEE Computer Society Journals}

\IEEEtitleabstractindextext{ \begin{abstract} Existing dimensionality reduction
methods are adept at revealing hidden underlying manifolds arising from
high-dimensional data and thereby producing a low-dimensional representation.
However, the smoothness of the manifolds produced by classic techniques over sparse and noisy data is not guaranteed. In fact, the embedding generated using such data may distort the geometry of the manifold and thereby produce an unfaithful embedding.  Herein, we propose a framework for
nonlinear dimensionality reduction that generates a manifold in terms of smooth
geodesics that is designed to treat problems in which manifold measurements are either sparse or corrupted by noise.  Our method generates a network structure for given
high-dimensional data using a nearest neighbors search and then produces piecewise
linear shortest paths that are defined as geodesics. Then, we fit points in each
geodesic by a smoothing spline to emphasize the smoothness. The robustness of
this approach for sparse and noisy datasets is demonstrated by the
implementation of the method on synthetic and real-world datasets.
\end{abstract}

}

\maketitle

\IEEEdisplaynontitleabstractindextext
\IEEEpeerreviewmaketitle

\ifCLASSOPTIONcompsoc
\IEEEraisesectionheading{\section{Introduction}\label{sec:introduction}}
\else
\section{Introduction}
\label{sec:introduction}
\fi

\IEEEPARstart{A}{dvanced} data collection techniques in today's world require
researchers to work with large volumes of nonlinear data, such as global climate
patterns \cite{aksoy2010pattern, yu2015big}, satellite signals
\cite{manjunath1996texture, jin2014classification},  social and mobile networks
\cite{chaker2017social, parwez2017big}, the human genome
\cite{liew2005pattern, berg2009large}, and patterns in collective
motion \cite{kong2014interactive, solmaz2012identifying}. Studying,
analyzing, and predicting such large datasets is challenging, and many such
tasks might be implausible without the presence of Nonlinear Dimensionality
Reduction (NDR) techniques. NDR interprets high-dimensional data using a reduced
dimension that corresponds to the intrinsic nonlinear
dimensionality of the data \cite{van2009dimensionality}. Manifolds are often
thought of as being smooth, however many existing NDR methods do not directly
leverage this important feature. Sometimes, ignoring the underlying smoothness
of the manifold can lead to inaccurate embeddings, especially when the data is \emph{sparse} or has
been contaminated by \emph{noise}.

Many NDR methods have been developed over the last two decades due to the lack
of accuracy and applicability of classic Linear Dimensionality Reduction (LDR)
methods such as Principal Component Analysis (PCA) \cite{jolliffe2002principal},
which finds directions of maximum variance, or Multi-Dimensional Scaling (MDS)
\cite{cox2000multidimensional}, which attempts to preserve the squared Euclidean
distance between pairs of points.  As the Euclidean distance used in MDS
to quantify the distance between points in the high-dimensional space rather than the
actual distance on the manifold, MDS has difficulties of inferring a faithful
low-dimensional embedding of non-linear data. The NDR method Isometric Mapping (Isomap), replaces the Euclidean metric in MDS with \emph{geodesic metric} to represent pairwise distances between points, successfully resolves the aforesaid problem in MDS \cite{tenenbaum2000global}. Although Isomap has been used to
analyze low-dimensional embedding of data from several domains, such as
collective motion \cite{delellis2014collective}, face recognition
\cite{yang2002face}, and hand-writing digit classification
\cite{yang2002extended}, this method can suffer from short-circuiting
\cite{balasubramanian2002isomap}, low-density of the data
\cite{lee2004nonlinear}, and non-convexity \cite{zha2007continuum}, all of
which can be magnified in the presence of noise.  It is therefore
our goal here to propose a new method which ameliorates some of
these issues as compared to Isomap.

\begin{figure}[htp]
    \centering
    \begin{subfigure}{0.24\textwidth}
        \centering
        \includegraphics[width= 2.2in]{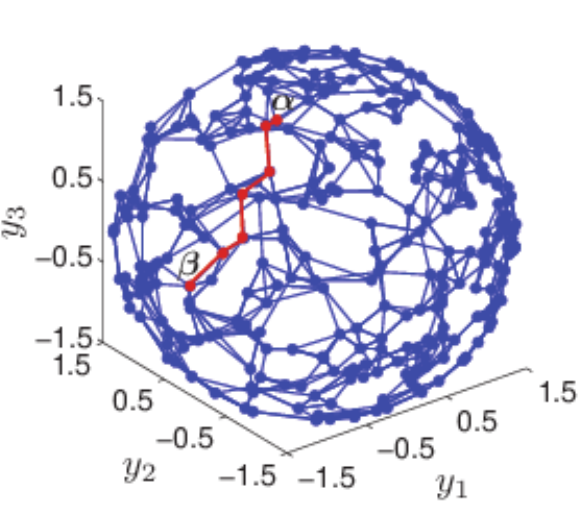}
        \caption{}
    \end{subfigure}%
    ~
    \begin{subfigure}{0.24\textwidth}
        \centering
	\vspace{2mm}
        	\includegraphics[height= 1.7in]{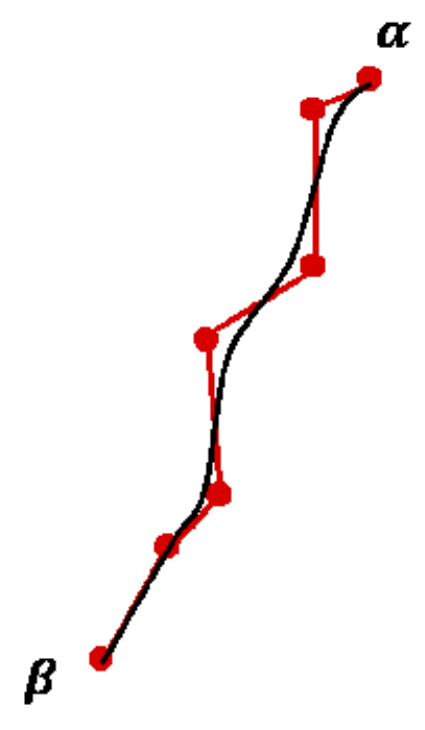}
	\vspace{4mm}
        \caption{}
    \end{subfigure}
    \caption{This figure demonstrates the lack of smoothness of the geodesics generated by Isomap. (a) Three nearest neighbors for each point (blue dots) of a spherical dataset of 300 points are found and joined by line segments (shown in blue) to create a graph structure.  The Isomap manifold distance between two arbitrary points $\bs{\alpha}$ and $\bs{\beta}$ is estimated as the length of the geodesic (red path), that is defined as the shortest path between two points, computed by using, for example, Floyd's algorithm \cite{floyd1962algorithm}. (b) However, our approach creates a \emph{smoothing spline}, shown by the black curve, that is fitted through the points in the geodesic as a better approximation of the distances on the smooth manifold than the geodesic distance.}
\label{fig:sphere_with_network}
\end{figure}

Generally, NDR approaches reveal smooth low-dimensional and nonlinear
manifold representations of high-dimensional data. While there are many unique
capabilities provided by current NDR methods, most of them encounter poor
performance in specific instances. In particular, many current NDR methods are not
adept at preserving the \emph{smoothness} of the embedded manifold when the data is sparse or noisy. Isomap closely mimics the underlying manifold's geometry
using a graph structure that it makes using a nearest neighbors search
\cite{friedman1977algorithm, agarwal1999geometric}, over the high-dimensional
data. The geodesic between two points on the manifold is defined as the shortest path in this graph between those two points. Geodesics are generally \emph{piecewise linear}, thus, the manifold constructed using geodesics in this method is not actually smooth at each node, as demonstrated in
Fig.~\ref{fig:sphere_with_network}(a). Moreover, given a sufficiently smooth manifold, the Isomap generated geodesic distance
will generally be an \emph{over-estimate} of the true manifold distance as demonstrated in Fig.~\ref{fig:sphere_with_network}(b). Of course, such issues are intensive in the presence of sparse and noisy measurements. Accordingly, herein we propose to replace the piecewise linear Isomap geodesic by a \emph{smoothing spline} as shown by the black curve in
Fig.~\ref{fig:sphere_with_network}(b) and \emph{consider the length of the
spline as the estimation of the manifold distance between points}.

There are few NDR methods found in the literature that utilize smoothing splines
for embedding like our approach.  For example, Local Spline Embedding (LSE) also
uses smoothing splines to perform the embedding \cite{xiang2009nonlinear}.
This method minimizes the reconstruction error of the objective
function and embeds the data using smoothing splines that map local coordinates
of the underlying manifold to global coordinates. Specifically, LSE assumes the
existence of a smooth low-dimensional underlying manifold and the embedding is
based on an eigenvalue decomposition that is used to project the data onto a
tangent plane. However, differing from our approach, LSE assumes that the data
is noise free and unaffected by anomalies. Another disadvantage of LSE is that it embeds the data into a space where the distances in this space are not faithful to the distances on the manifold.
The Principal Manifold Finding Algorithm (PMFA) is another NDR method that also uses cubic smoothing splines to
represent the manifold and then quantifies the intrinsic distances of the points
on the manifold as lengths of the splines \cite{gajamannage2015dimensionality}.
However, this approach embeds high-dimensional data by reducing the
reconstruction error over a two-dimensional space. As this method only performs
two-dimensional embeddings, its applicability is limited for problems with
large intrinsic dimensionality. As we will demonstrate in Sec.~\ref{sec:per_analysis}, our
proposed method overcomes the limitations of the methods LSE and PMFA.

This paper is structured as follows.  In Sec.~\ref{sec:mds_isomap}, we will
detail the MDS and Isomap algorithms and describe the evolution of our NDR
method from these methods. Sec.~\ref{sec:sge} presents our NDR method, Smooth
Geodesic Embedding (SGE), that fits geodesics, as in Isomap, by smoothing
splines. We analyze the performance of the SGE method in Sec.~\ref{sec:per_analysis} versus three NDR methods, Isomap, LSE, and PMFA, using three representative examples: a
semi-spherical dataset, images of faces, and images of hand written digits.
Finally, we provide discussion and conclusions in Sec.~\ref{sec:conclusion}.

\section{Multidimensional scaling and Isomap}\label{sec:mds_isomap}
We begin our analysis by deriving the mathematical details of the LDR method MDS. Then, we proceed to discuss Isomap which replaces the Euclidean distance in MDS by a geodesic distance. Next, we derive our method, SGE, as an extension of Isomap that fits geodesics by smoothing splines.

\subsection{Multidimensional scaling}\label{sec:mds}
Multidimensional scaling is a classic LDR algorithm that leverages the squared Euclidean distance matrix $\bs{D}^2=[d_{ij}^2]_{n\times n}$; where $d_{ij} = \|\bs{y}_i - \bs{y}_j\|_2$ and $n$ is number of points in the data. Here, $\bs{y}_i$, $\bs{y}_j\in \mathbb{R}^{n\times 1}$, are two points on the high-dimensional dataset $\bs{Y}=[\bs{y}_1; \dots; \bs{y}_i; \dots; \bs{y}_j; \dots; \bs{y}_n]$. This method first transforms the squared distance matrix $\bs{D}^2$ into a Gram matrix $\bs{S}=[s_{ij}]_{n\times n}$, which is derived by \emph{double-centering} \cite{lee2007nonlinearb} the data using
\begin{equation}\label{eqn:double_centering}
s_{ij}=-\frac{1}{2}\big[d^2_{ij}-\mu_i(d^2_{ij}) -\mu_j(d^2_{ij})+\mu_{ij}(d^2_{ij})\big].
\end{equation}
Here, $\mu_i(d^2_{ij})$ and $\mu_j(d^2_{ij})$ are the means of the $i$-th row and $j$-th column, respectively, of the squared distance matrix, and $\mu_{ij}(d^2_{ij})$ is the mean of the entire matrix $D^2$. MDS then computes the Eigenvalue Decomposition (EVD) of $\bs{S}$ as
\begin{equation}\label{eqn:evd}
\bs{S}=\bs{U}\bs{\Sigma}\bs{U}^T,
\end{equation}
where $\bs{U}$ is a unitary matrix ($\bs{U}^T\bs{U}=\bs{I}$) providing the eigenvectors $\bs{U}^T$ and a diagonal matrix of eigenvalues $\bs{\Sigma}$. The Gram matrix $\bs{S}$, that is made from the squared Euclidean distance matrix $\bs{D}^2$, is symmetric and Semi-Positive Definite (SPD)\footnote{A symmetric $n\times n$ matrix $\bs{M}$ is said to be SPD, if $\bs{z}^T \bs{M} \bs{z}\ge 0$ for all non-zero $\bs{z}\in\mathbb{R}^{n \times 1}$. \cite{lee2007nonlinearb}.}  Thus, all the eigenvalues of a SPD matrix $\bs{S}$, and both the Singular-Value Decomposition (SVD) and the EVD of $\bs{S}$ are the same~\cite{lee2007nonlinearb}. $\bs{\Sigma}$ and $\bs{U}^T$ are arranged such that the diagonal of $\bs{\Sigma}$ contains the eigenvalues of $S$ in descending order, and the columns of $\bs{U}^T$ represent the corresponding eigenvectors in the same order. We estimate $p$-dimensional latent variables of the high-dimensional dataset by
\begin{equation}\label{eqn:latent_var}
\hat{\bs{X}}=\bs{I}_{p\times n}\bs{\Sigma}^{1/2}\bs{U}^T.
\end{equation}
Here, $\bs{I}_{p\times n}$ is a matrix made from first $p$ rows of the identity matrix $\bs{I}_{n\times n}$ and $\hat{\bs{X}}$ is the $p$-dimensional embedding of the input data $\bs{Y}$.
However, due to both the approximations in our method and finite precision in computer arithmetic, the computed $\bs{S}$ might deviate slightly from being SPD. The EVD of such an $\bs{S}$ might have small negative eigenvalues and these negative eigenvalues would violate Eqn.~(\ref{eqn:latent_var}).  Accordingly, as we will discuss in Sec.~\ref{sec:isomap}, we replace the EVD on $\bs{S}$ by the SVD.
Multidimensional scaling has limited applicability as it is inherently a linear method. However, the NDR scheme Isomap overcomes this drawback by employing geodesic distance instead of Euclidean distance.

\subsection{Isomap}\label{sec:isomap}
Isomap creates a graph structure, based upon high-dimensional data, that estimates the intrinsic geometry of the manifold. The graph structure created by Isomap can be parameterized in multiple ways, but herein we
focus on the parameter $\delta$ which measures the number of \emph{nearest neighbors}
to a given point \cite{agarwal1999geometric}.  The nearest neighbor collection for each point is transformed into a graph structure by treating points as graph nodes and connecting each pair of nearest neighbors by an edge having the weight equal to the Euclidean distance between the two points. Given such a graph, the distance between any two points is measured as the \emph{shortest path distance in the graph}, which is commonly called the \emph{geodesic distance}.

The geodesic distance between any two points in the data can be computed in many ways, including Dijkstra's algorithm \cite{dijkstra1959note}, one that the original Isomap used. Dijkstra's algorithm, having computational complexity of $\mathcal{O}(n^2)$ when used for adjacency matrices, computes the shortest path between two pairs of points at a time \cite{ray2012graph}. Since our dataset has $n(n-1)/2$ distinct pairs of points (we make combinations of 2 points out of $n$ points), the total complexity of the  Dijkstra's algorithm is $\mathcal{O}(n^4/2)$, [$\mathcal{O}(n^3(n-1)/2) \approx \mathcal{O}(n^4/2)$]. However, Floyd's algorithm \cite{floyd1962algorithm} computes shortest paths between all pairs of points in one batch with the computational complexity of $\mathcal{O}(n^3)$ \cite{ray2012graph}, which is more efficient than utilizing Dijkstra's algorithm. Thus, we replace Dijkstra's algorithm in Isomap with Floyd's algorithm.

As in MDS, we first formulate the doubly centered matrix $\bs{S}$ from the squared geodesic distance matrix using Eqn.~(\ref{eqn:double_centering}). The doubly centered matrix here is not necessarily SPD as we \emph{approximate} the true geodesic distance matrix by the shortest graph distance \cite{lee2007nonlinearb}. In fact, our computational process uses several numerical approximations that might also cause $\bs{S}$ to deviate from being SPD. Thus, the eigenvalue decomposition of matrix $\bs{S}$ might produce negative eigenvalues and  Eqn.~(\ref{eqn:latent_var}) does not hold in this case. To overcome this problem, it is the standard to perform the SVD over the Gram matrix $\bs{S}$ as
\begin{equation}\label{eqn:svd}
\bs{S}=\bs{V}\bs{\Sigma} \bs{U}^T,
\end{equation}
where $\Sigma$ is a diagonal matrix of singular values (non negative), and $U$ and $V$ are unitary matrices. The $p$ latent variables of the higher dimensional input data are revealed by Eqn.~(\ref{eqn:latent_var}) with $\bs{\Sigma}$ and $\bs{U}^T$ obtained from Eqn.~(\ref{eqn:svd}).

Isomap emphasizes nonlinear features of the manifold, however the lengths measured using geodesics might not faithfully reflect the true manifold distance, as we demonstrate in Fig.~\ref{fig:sphere_with_network}.  Accordingly, we propose to  overcome this drawback in Isomap by utilizing a smoothing approach for geodesics.

\section{Smoothing geodesics embedding}\label{sec:sge}

Our goal is to fit the geodesics computed in Isomap with smoothing splines to closely mimic the manifold and preserve the geometry of the embedding. Classic smoothing spline constructions \cite{de1972calculating} require one input parameter, denoted by $s$, that controls the smoothness of the spline fitted through the points in a geodesic.  Our proposed method, SGE, has five parameters:
\begin{itemize}
\item $\delta$ (inherent from Isomap) controls the number of nearest neighbors,
\item$\mu_s$ controls the smoothness of the splines,
\item $\nu$ controls the threshold of the length of splines before reducing the order of the spline to the next level,
\item $h$ controls the number of discretizations that the method uses to evaluate the length of a spline, and
\item finally, $p$ prescribes the number of embedding dimensions (latent variables).
\end{itemize}
Note that, we will provide details of the parameters $\mu_s$, $\nu$, and $h$ later in this section.

Here, we demonstrate our approach by fitting a spline over an arbitrary geodesic $\mathcal{G}$, having $m\ge2$ points, in the graph created by a nearest neighbors search algorithm.  For an index $k$ we have that $d$-dimensional points in $\mathcal{G}$ are given by
\begin{equation}\label{eqn:geo_pts}
\big\{\bs{y}_k=[y_{1k}, \dots, y_{dk}]^T\vert k=1 \dots, m\big\}.
\end{equation}
For each dimension $l\in \big\{1, \dots, d\big\}$, we fit $\big\{y_{lk}\vert k=1 \dots, m\big\}$ using one dimensional smoothing splines $\hat{f}_l(z)$ of order $\theta+1$ that are parameterized in $z \in [0,1]$ by minimizing
\begin{equation}\label{eqn:spline}
\sum^m_{k=1}\big[y_{lk}-\hat{f}_l(z_k)\big]^2 + s\int^1_0 \big[\hat{f}_l^{(\theta)}(z)\big]^2dz
\end{equation}
as in \cite{de1972calculating}. Here, $(\theta)$ represents the order of the derivative of $\hat{f}_l$, and $z_k$ is a discretization of the interval $[0,1]$ such that $z_1=0$, $z_k=(k-1)/(m-1)$, and $z_m=1$. Minimizing of Eqn.~(\ref{eqn:spline}) yields $d$ one-dimensional smoothing splines $\{\hat{f}_l(z)|l=1,\dots,d\}$. We combine these one dimensional splines and obtain a $d$-dimensional smoothing spline of the points $\big\{\bs{y}_k\vert k=1 \dots, m\big\}$ in $\mathcal{G}$ as,
\begin{equation}\label{eqn:dspline}
\hat{\bs{f}}(z)=[\hat{f}_1(z), \dots, \hat{f}_d(z)]^T,
\end{equation}
which is called the \emph{smooth geodesic}. In numerical implementations, the order $\theta+1$ of the spline $\hat{f}$ should be less than number of points $m$ in the geodesic \cite{de1972calculating}.

Choosing the order of a spline $\theta$ is challenging, since while a spline with some specified order might perfectly fits the data, another spline with a different order might weakly fits the data. The length of the fitted spline between two points is defined as the manifold distance between those two points, thus an over-fitted spline might provide an incorrect manifold distance. To overcome this problem, we introduce the spline threshold $\nu$ (in percentage) which allows the maximum length of a spline that can yield beyond the length of the corresponding geodesic. If the length of a spline with a specific order exceeds this limit, SGE keeps on reducing the order of the spline by one unit until the length of the new spline satisfies the threshold. If non of the orders satisfy the threshold, then SGE assumes the manifold distance is the default distance which is defined to be the geodesic distance. We opt for this procedure, as it is worthwhile to fit a spline with a lower order when a higher order spline fails numerically. Choosing the order of the smoothing spline can also be considered as a trial and error process. For the simplicity, we choose the order here by using the threshold percentage $\nu$. Again for the simplicity, we start by fitting a cubic smoothing spline over the points on a given geodesic and then reduce the order if the length doesn't meet the threshold. Cubic smoothing splines emphasize smoothness while involving a low fitting error. We empirically observed that over-fitting occurs very rarely in SGE, thus most of the geodesics were fitted with cubic smoothing splines.

Bellow we present our procedure of choosing the order of a spline, fitting points $\big\{\bs{y}_k\vert k=1 \dots, m\big\}$ on a geodesic, under three main cases (1, 2, and 3) and some sub-cases (a, b, \dots):

\begin{itemize}
\item \textbf{Case--1} If $m\ge4$:
\begin{itemize}
\item \textbf{Case--a:} we first fit the points in the geodesic with a cubic smoothing spline $\hat{\bs{f}}(z)$ where $z\in[0,1]$ according to Eqn.~(\ref{eqn:spline}) and  Eqn.~(\ref{eqn:dspline}). Note that, a cubic smoothing spline is represented by $\theta=2$  in Eqn.~(\ref{eqn:spline}). We discretize this spline into $h$ segments $z_{k_1}=(k_1-1)/(h-1); k_1=1,\dots, h$ and compute the length,
\begin{equation}\label{eqn:len_spl}
d_{\hat{\bs{f}}}=\sum^{h-1}_{k_1=1}\|\hat{\bs{f}}(z_{k_1+1})-\hat{\bs{f}}(z_{k_1})\|.
\end{equation}
 Then, the length $d_{\hat{\bs{f}}}$ is compared with the corresponding geodesic distance
\begin{equation}\label{eqn:len_geo}
d_\mathcal{G}=\sum_{k=1}^{m-1}\|\bs{y}_{k+1}-\bs{y}_k\|.
\end{equation}
If $d_{\hat{\bs{f}}}<d_\mathcal{G}(100+\nu)/100$ (so that $\nu$ is thought of as a percentage), then we accept $d_{\hat{\bs{f}}}$ as the length of the smooth geodesic, otherwise we proceed to Case--b. The parameter $\nu$ (in percentage) defines the threshold (the upper bound) of the length of the spline $\hat{\bs{f}}$ that is allowed to exceed from the length of the corresponding geodesic.\\
\item \textbf{Case--b:} we fit the data with a quadratic (i.e., $\theta=1$ ) spline $\hat{\bs{f}}$ according to Eqn.~(\ref{eqn:spline}) and  Eqn.~(\ref{eqn:dspline}) and compute the length of the quadratic spline using Eqn.~(\ref{eqn:len_spl}). If $d_{\hat{\bs{f}}}<d_\mathcal{G}(100+\nu)/100$, then we accept $d_{\hat{\bs{f}}}$ as the  length of the smooth geodesic, otherwise proceed to Case--c. \\
\item \textbf{Case--c:} we make a linear (i.e., $\theta=0$) fit $\hat{\bs{f}}$ according to Eqn.~(\ref{eqn:spline}) and Eqn.~(\ref{eqn:dspline}), and measure the length using Eqn.~(\ref{eqn:len_spl}). If $d_{\hat{\bs{f}}}<d_\mathcal{G}(100+\nu)/100$ in the linear fit, then we accept $d_{\hat{\bs{f}}}$, otherwise proceed to Case--d. \\
\item\textbf{Case--d:} instead of fitting a spline, we assume the original geodesic itself as the fit and treat $d_\mathcal{G}$ as the length of the smooth geodesic.
\end{itemize}
\item \textbf{Case--2} If $m=3$:\\
The spline fitting process here is started from fitting a quadratic spline as only three points are in the geodesic. Thus, we carry-out all the Cases b--d as in Case--1.
\item \textbf{Case--3} If $m=2$:\\
We have only two points in the geodesic, thus, we perform Cases c--d as in Case--1.
\end{itemize}

We use the smoothing parameter $s$ to offset the spline fit between no fitting error (when $s=0$) and the best smoothness (when $s\rightarrow \infty$). The parameter $s$ controls the sum of square errors between the training points and the fitted function. The best value for $s$ ensuring the least error while having a sufficient smoothness is bounded by a function of the number of points in the geodesic as
\begin{equation}\label{eqn:spl_interval}
m-\sqrt{m}\le s \le m+\sqrt{m},
\end{equation}
\cite{reinsch1967smoothing}. Since the number of points in geodesics vary, we are unable to input a one-time value as the smoothing parameter into the method that satisfies the inequality (\ref{eqn:spl_interval}). In order to control this, here we introduce a new parameter called the smoothing multiplier $\mu_s\ge0$ such that $s=\mu_s m$. Thus, for  input parameter $\mu_s$, such that
\begin{equation}\label{eqn:mus_interval}
1-1/\sqrt{m}\le \mu_s \le 1+1/\sqrt{m},
\end{equation}
SGE uses different smoothing levels for different splines based on number of points on the geodesics ($m$).

For each pair of point in the dataset, say they are indexed by $i$ and $j$, we execute the aforesaid procedure and approximate the length of the smooth geodesic $d_{ij}$.  Then, we square the entries $d_{ij}$ and create the matrix $\bs{D}^2=[d^2_{ij}]_{n\times n}$. We perform double centering on $\bs{D}^2$ using Eqn.~(\ref{eqn:double_centering}) to obtain the doubly centered matrix $\bs{S}$. Then, we compute SVD as in Eqn.~(\ref{eqn:svd}) followed by computing $p$-dimensional latent variables $\hat{\bs{X}}$ according to Eqn.~(\ref{eqn:latent_var}). A summary of the SGE method is presented in Algorithm \ref{alg:algorithm}.

\begin{algorithm*}[!htp]
\caption{ \textit{Smooth Geodesics Embedding (SGE).
\\ Inputs: Data ($\bs{Y}$), number of nearest neighbors ($\delta$), smoothing multiplier ($\mu_s$), spline threshold percentage ($\nu$), number of discretizations ($h$), and embedding dimensions ($p$).
\\Outputs: List of $p$ largest singular values ($\lambda_l;l=1,\dots,p$) and $p$-dimensional embedding ($\hat{\bs{X}}$). }}\label{alg:detecting_transitions}.
\begin{algorithmic}[1]
\State For each point in $\bs{Y}$, choose $\delta$ number of nearest points as neighbors \cite{friedman1977algorithm}.
\State Consider all the point in $\bs{Y}$ as nodes and if any two nodes are chosen to be neighbors in 1, then join them  by an edge having the length equal to the Euclidean distance between them. This step converts the dataset into a graph.
\State For each pair of nodes in the graph, find the points $\mathcal{G}=\big\{\bs{y}_k\vert k=1 \dots, m\big\}$ in the shortest path using the Floyd's algorithm \cite{floyd1962algorithm}. Here, $m=|\mathcal{G}|\ge2$.
\State The points in $\mathcal{G}$ are fitted with a smoothing spline and its length is computed:
\newline
Case--1 ($m\ge4$):
\begin{addmargin}[1em]{2em}
Case--a:
\begin{addmargin}[1em]{2em}
Fit $\mathcal{G}$ with a cubic smoothing spline using Eqn.~(\ref{eqn:spline}) and Eqn.~(\ref{eqn:dspline}), then approximate the length $d_{\hat{\bs{f}}}$ of the spline using Eqn.~(\ref{eqn:len_spl}). Let, the length of the geodesic is $d_\mathcal{G}$ [Eqn.~(\ref{eqn:len_geo})].  If $d_{\hat{\bs{f}}}<d_\mathcal{G}(100+\nu)/100$, then accept $d_{\hat{\bs{f}}}$ as the length of the smooth geodesic, otherwise proceed to Case--b.
\end{addmargin}
Case--b:
\begin{addmargin}[1em]{2em}
Fit $\mathcal{G}$ with a quadratic smoothing spline using Eqn.~(\ref{eqn:spline}) and Eqn.~(\ref{eqn:dspline}).  Approximate the length $d_{\hat{\bs{f}}}$ of the spline using Eqn.~(\ref{eqn:len_spl}). If $d_{\hat{\bs{f}}}<d_\mathcal{G}(100+\nu)/100$, then accept $d_{\hat{\bs{f}}}$ as the length of the smooth geodesic, otherwise proceed to Case--c.
\end{addmargin}
Case--c:
\begin{addmargin}[1em]{2em}
Fit $\mathcal{G}$ with a linear smoothing spline using Eqn.~(\ref{eqn:spline}) and Eqn.~(\ref{eqn:dspline}). Approximate the length $d_{\hat{\bs{f}}}$ of the spline using Eqn.~(\ref{eqn:len_spl}). If $d_{\hat{\bs{f}}}<d_\mathcal{G}(100+\nu)/100$, then accept $d_{\hat{\bs{f}}}$ as the length of the smooth geodesic, otherwise proceed to Case--d. 
\end{addmargin}
\end{addmargin}
\end{algorithmic}\label{alg:algorithm}
\end{algorithm*}

\begin{algorithm*}[!htp]
\begin{algorithmic}[1]
\Statex 
\begin{addmargin}[1em]{2em}
Case--d:
\begin{addmargin}[1em]{2em}
Consider $d_\mathcal{G}$ as the approximated length of the smooth geodesic.
\end{addmargin}
\end{addmargin}
Case--2 ($m=3$):
Perform Cases b--d similarly as in Case--1.
\newline
Case--3 ($m=2$):
Perform Cases c--d similarly as in Case--1.
\setcounter{ALG@line}{4}
\State Fill the distance matrix $\bs{D}^2=[d^2_{ij}]_{n\times n}$ where $d_{ij}$ is the length of the smooth geodesic between nodes $i$ and $j$ computed in 3-4. Double center $\bs{D}^2$ and convert it to a Gramian matrix $\bs{S}$ using the Eqn.~(\ref{eqn:double_centering}).
\State Perform the SVD on $\bs{S}$ using Eqn.~(\ref{eqn:svd}) and extract $p$ largest singular values $\lambda_l;l=1,\dots,p$ along with the latent variable $\hat{\bs{X}}$ as given by Eqn.~(\ref{eqn:latent_var}).
\end{algorithmic}
\end{algorithm*}

\begin{figure}[htp]
	\vspace{10pt}
        	\centering
        	\includegraphics[width=3.5in]{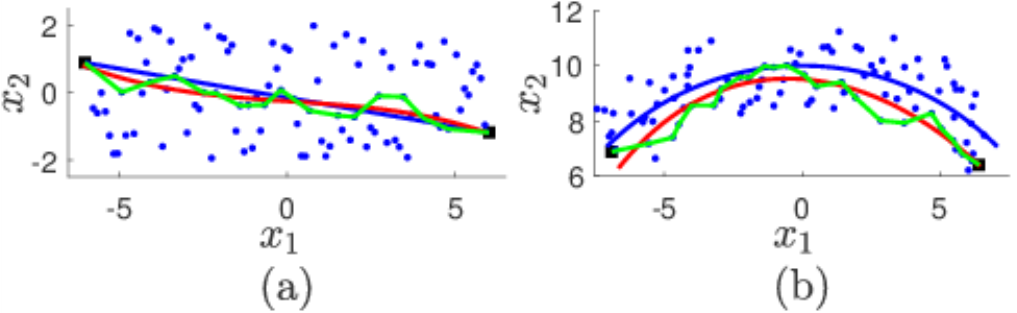}
        	\caption{Trade-offs of the shortest paths and smoothing splines from the true geodesics. (a) The first dataset (blue points) which is noise free is sampled from a two-dimensional rectangular shaped manifold. (b) The second dataset (blue points) is sampled from a one-dimensional manifold (blue curve) representing an arc after imposing a uniform noise. We run nearest neighbor search algorithm over both datasets with four nearest neighbors ($\delta=4$) and create a graph structure in each dataset. Then, we compute the geodesics (green curves) between two points (black squares) in the datasets, and then fit the points on each geodesic using a cubic smoothing splines (red curves) with $\mu_s=1$.  Note that, the blue curves represent the true manifold distance between two black squares.  In (a) we see that the smoothing spline more faithfully representing the true geodesic distance of on the noise free manifold, while in (b) we see that both SGE and Isomap can suffer in the presence of noise.}
	\label{fig:spl_approximation}
\end{figure}

Approximate geodesics arising from graph shortest paths in a finite dataset are different than the true geodesics. However, smoothing splines that fit points on geodesics are capable of closely approximating the true geodesics of finite, sparse, and noisy datasets sampled from a manifold, as shown in Fig.~\ref{fig:spl_approximation}.

The smoothing spline approach in SGE approximates true geodesic distance of sparse samples of data more accurately than that of the graph distance used in Isomap [Fig.~\ref{fig:spl_approximation}(a)]. Note, the shortest path between two points on a noise free manifold converges to the true geodesic of the manifold as the number of sample points approaches infinity \cite{bernstein2000graph}. Thus, Isomap can convergently approximate the manifold distances using shortest graph distances and makes better predictions with dense samples of data. However, as SGE fits vertices on shortest paths with smoothing splines, we demonstrate herein that SGE converges to the true manifold distance at a faster rate than that of Isomap.  In particular, our smoothing approach assures better predictions than that of Isomap even under sparse samples of data as we justify in Sec.~\ref{sec:per_analysis}.

However, both the smoothing spline approximation of noisy data in SGE and the geodesic approximation of noisy data in Isomap, might not faithfully represent the real manifold [Fig.~\ref{fig:spl_approximation}(b)]. This is because the Floyd's algorithm might find a shortest path that is different than the true manifold if the data is contaminated with noise. Since smoothing splines fit the data points on such shortest paths, they might also deviated from the true manifold. Thus, both geodesics in Isomap and smoothing splines in SGE demonstrate lack of convergence to the true manifold even at the limit of infinite samples of noisy data. We believe that we can produce a convergent version of the SGE method, if we first compute the curvature of the manifold and then utilize a technique that helps to choose the shortest paths close to the manifold. We will explain this technique with details as a future work in Sec.~\ref{sec:conclusion}.  However, we demonstrate that, empirically speaking, the smoothing spline approach is a better replacement for graph shortest paths even when the data is contaminated with noise using the examples in Sec.~\ref{sec:per_analysis} .

Note that, we provide more examples in Sec.~\ref{sec:per_analysis} that support
the accuracy of embedding noisy datasets versus that of sparse datasets. The
reason for that is, convergence of both methods, Isomap and SGE, of embedding
sparse datasets is supported by \cite{bernstein2000graph}, while the convergence
of both methods is not guaranteed when the data is contaminated with noise.

\section{Performance analysis}\label{sec:per_analysis}
In this section, we demonstrate the effectiveness SGE, versus Isomap and PMFA,  in example 1 and then that of SGE versus Isomap in examples 2 and 3. PMFA first makes $n^1_c$ non overlapping slices of data along the direction of the largest singular vector and then makes $n^2_c$ slices along the second largest singular vector. Then, the data in each slice is fitted with a cubic smoothing spline of smoothing parameter $s_r$. The user input parameters in this method are $n^1_c$, $n^2_c$, and $s_r$ . The grid structure represented by all the cubic smoothing splines  is used as the local intrinsic coordinate system that we use to measure the embedding distances. Large $n^1_c$ and $n^2_c$ values make thin slices with few points. Accordingly, the low density of points in such a slice can cause cubic smoothing splines to weakly fit the data and misinterpret the manifold. In contrast, small $n^1_c$ and $n^2_c$ make few slices and create a sparse grid structure. This sparse grid structure might loose the geometry of the manifold.  Accordingly, we use moderate values for the parameters $n^1_c$ and $n^2_c$, say 10 each, in the examples we provide. While big values of $s_r$ make less oscillatory cubic smoothing splines, small values can make highly oscillatory cubic smoothing splines. As stated in \cite{gajamannage2015dimensionality}, the best value for $s_r$ is 0.9, and we use that value in the numerical examples in this manuscript.

LSE is also a NDR method that employs splines approach in their embedding routing.  This method requires one input parameter for number of nearest neighbors ($\delta$) and it projects $\delta$ neighbors of each points into a tangent space with local coordinates. Each such local coordinate is then mapped to its own single global coordinate with respect to the underlying manifold using splines. The parameter $\delta$ in LSE is also a input parameter in SGE, thus, details for choosing the value for this parameter will be explained later when the parameter values of SGE are explained.

For all the examples in this section, we set $\nu=10\%$ and $h=100$ in SGE. Setting $\nu$ to a high value increases the  tendency of fitting points on the geodesics in SGE with cubic smoothing splines than fitting those points with splines having order less than three. SGE is fabricated to reveal a smooth underlying manifold that is ensured by cubic smoothing splines than a spline with a low order. However, high $\nu$ values sometimes over-fit the data and that will then yield inaccurate embedding. We empirically learned that  setting $\nu$ to $10\%$ can exclude both of aforesaid extremes. Each spline is discretized to $h$ segments and the length of the spline is computed as sum of linear lengths of these segments. While a big $h$ gives a very accurate length for the spline, it increases the computational time as SGE has to compute lengths of $n(n-1)/2$ splines for a dataset of $n$ points. Thus, we set $h=100$ since the accuracy of the spline lengths by 100 discretization is satisfactory for our study. 

We set $\delta=3$ or $4$, and $\mu_s=1$ in SGE, if not stated otherwise. Each point in the dataset is adjacent to $\delta$ number of nearest neighbors and the graph structure is made. Setting a big value for $\delta$ will create more edges in the graph and that might loose the topology of the graph as geodesics in this case might not infer the true curvature of the manifold. However, a small value of $\delta$ might produce multiple connected components in the graph where SGE treats the large connected component and neglects others in this case. We set $\delta=3$ or $4$, since we empirically observed that these values stay in the middle of aforesaid extremes. Choosing a best value for $\mu_s$ is challenging, thus based on Eqn.~\eqref{eqn:mus_interval}, we set $\mu_s=1$, since this value provides both a perfect smoothness and a better fit for the splines.

\begin{figure*}[htp]
	\vspace{10pt}
        	\centering
        	\includegraphics[width=4.8in]{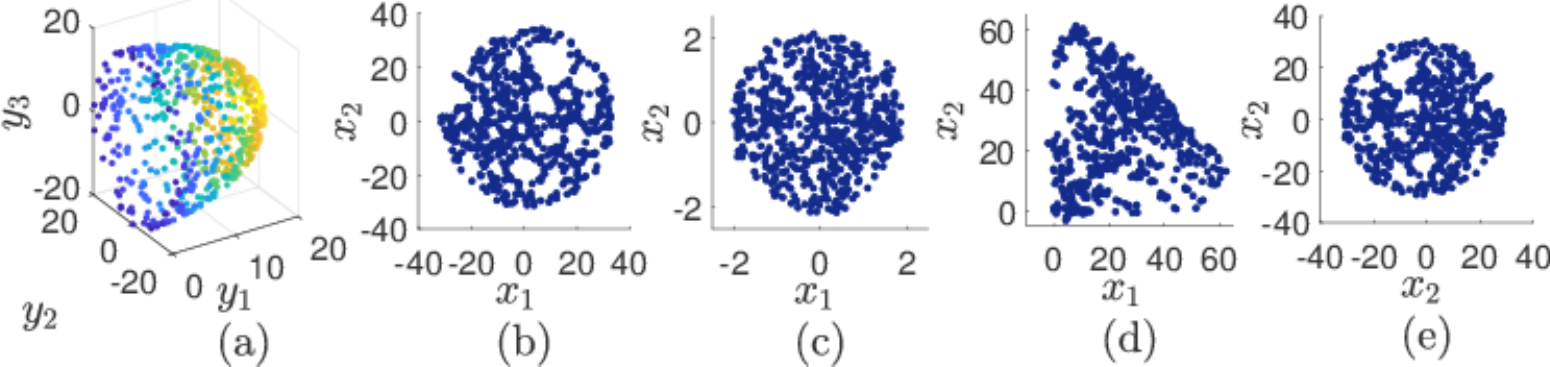}
        	\caption{Two dimensional embedding of (a) a noise free semi-sphere of 600 points using (b) Isomap, (c) LSE, (d) PMFA, and (e) SGE.}
	\label{fig:emb4methods}
\end{figure*}

The NDR methods that we utilize in this section should preserve the pairwise distances between data and embedding in order to compare$\backslash$contrast them using two distance metrics [Eqn.~(\ref{eqn:mad}) and Eqn.~(\ref{eqn:adj_error})] that we use to compute the embedding error in this paper. To visualize an instance of embedding  of these four methods SGE, Isomap, PMFA, and LSE, we embed a dataset sampled from a semi-sphere of 600 points [Fig.~\ref{fig:emb4methods}(a)] defined by
\begin{align}\label{eqn:sphere}\nonumber
y_1 &= r\cos(\gamma_1)\cos(\gamma_2), \\
y_2 & = r\cos(\gamma_1)\sin(\gamma_2), \\ \nonumber
y_3 & = r\sin(\gamma_1),\\ \nonumber
\end{align}
for $\gamma_1=\mathcal{U}[-\pi/2, \pi/2]$ and $\gamma_2=\mathcal{U}[0, \pi]$, where $\mathcal{U}[a,b]$ denotes a uniform distribution between $a$ and $b$. Here, $r$ is the radius of the semi-sphere which is set to $20+\mathcal{N}[0,\eta^2]$, where $\mathcal{N}[0,\eta^2]$ is a random variable sampled from a Gaussian distribution with mean 0 and variance $\eta^2$. We set $\eta=0$ as we need this semi-sphere to be noise free. 

We compute two dimensional embedding of this semi-sphere [Fig.~\ref{fig:emb4methods}(b--e)] using Isomap with $\delta=3$; LSE with $\delta=3$; PMFA with $n^1_c = n^2_c = 10$ and $s_r=0.9$; and SGE with $\delta=3$, $\mu_s=1$, $\nu=10\%$, and $h=100$. According to Fig.~\ref{fig:emb4methods}, moving from data to embedding, LSE shrinks the distances in the embedding while others seem preserve the original distance between points. Thus, we omit LSE from this analysis and only rely on the rest of the methods since two distance preserving error metrics that we used here can't be implemented for LSE. Moreover, PMFA is computationally expensive when the data is high dimensional (as stated in \cite{gajamannage2015dimensionality}) like in face images of example 2 where the dataset is 4096 dimensions, and hand written digits of example 3 where the dataset is 784 dimensions. Thus, we omit the implementation of PMFA for the datsets in those two examples.

As the first example, we use a synthetic three dimensional dataset of a semi-sphere to analyze the performance of SGE with respect to neighborhood size ($\delta$), smoothness ($\mu_s$), sparsity ($n$), and noise ($\eta$). Then, we study the performance of SGE using two high-dimensional standard benchmark datasets: 1) face images \cite{deeplearning}; and 2) images of handwritten digits (2's, 4's, 6's, and 8's) \cite{lecun2016the}. We analyze the performance of SGE versus Isomap and PMFA in example 1, and that of SGE versus Isomap in examples 2 and 3.

\subsection{Embedding of a semi-sphere}\label{sec:semi-sphere}
We begin this section by embedding a synthetic dataset, sampled from a semi-spherical manifold, using SGE, Isomap,  and PMFA to demonstrate the key concepts of our proposed SGE technique since, in this case, we can analytically compute the manifold distance on the semi-sphere and then compare it with the embedding distances computed by above NDR methods. 

First, we compare the performance of SGE with changing $\delta$ and $\mu_s$ by embedding a sample 600 points generated from the manifold defined by Eqn.~\ref{eqn:sphere} with $\eta=2$. Both $\mu_s$ in SGE and $s_r$ in PMFA are parameters to control the smoothness of the splines, however, while $\mu_s$ can only take any non negative value, chosen $s_r$ should be in $[0,1]$ \cite{gajamannage2015dimensionality}. Since we are unable to compare SGE and PMFA in this context, we only provide here a comparison between Isomap and SGE.

We set the spline threshold $\nu$ and spline discretization $h$ to be $10\%$ and 100, respectively, and run the SGE algorithm repeatedly over the spherical dataset with $\delta=2, 3, \dots, 8$; $\mu_s=0, 0.1, \dots, 1.0$ and obtain two-dimensional embeddings. Here, we have 77 different pairs of $\delta$'s and $\mu_s$'s, those then produce 77 two-dimensional embeddings. Now, we asses the performance of the methods in terms of distance preserving ability between the original data and the embedding. For each such embedding (77 in total), we compute distances  between points in the embedding space using the Euclidean distance metric that we denote by $\bs{D}_S$. Now, we run Isomap with the same sequence of $\delta$'s above and obtain its two-dimensional embeddings. The Euclidean distance matrix for the embedding of Isomap is denoted by $\bs{D}_I$. Now, we compute the true manifold distances between points of the dataset using the cosine law. If $\bs{\alpha}$ and $\bs{\beta}$ are two points on a semi-sphere with radius $r$, the manifold distance $d$ is given by
\begin{equation}\label{eqn:mani_dist}
d = r\gamma; \ \gamma=\cos^{-1}\begin{pmatrix}\frac{\bs{\alpha}\bs{\beta}}{|\bs{\alpha}||\bs{\beta}|}\end{pmatrix},
\end{equation}
\cite{stewart2012essential}. We compute all the pairwise distances using Eqn.~(\ref{eqn:mani_dist}) and form the distance matrix $\bs{D}_M$ of the data sampled on the manifold.

\begin{figure}[htp]
        	\centering
        	\includegraphics[width=3.5in]{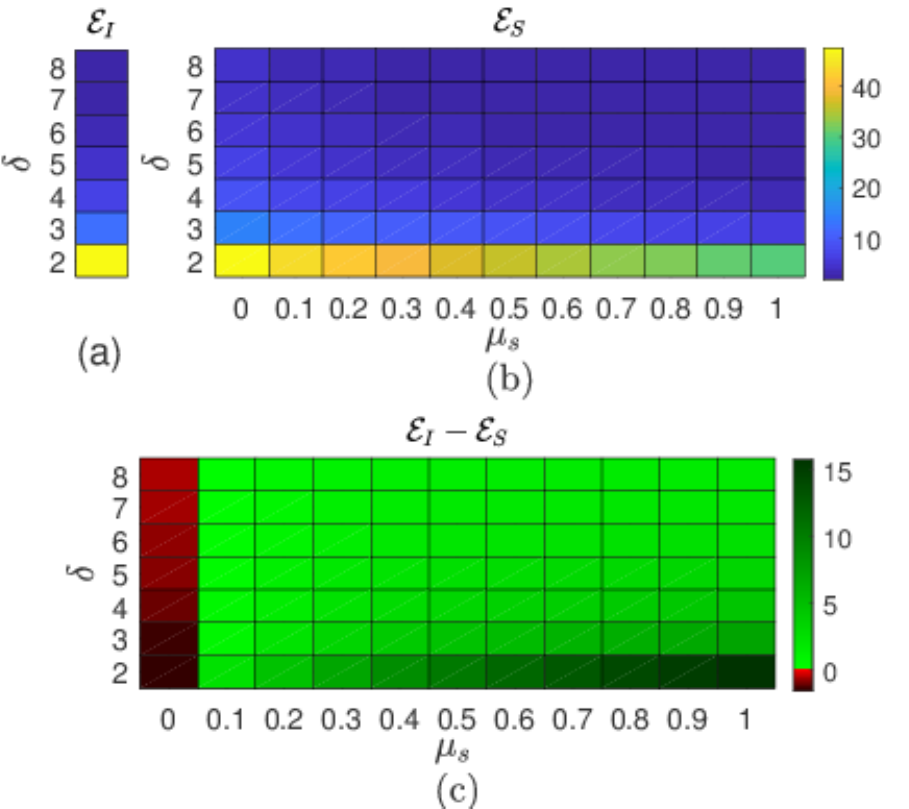}
        	\caption{Analyzing the performance of Isomap and SGE embeddings using Mean Absolute Deviation (MAD). Herein, we compute, (a) MAD between Isomap embedding and data, denoted by $\mathcal{E}_I$, for different neighborhood sizes ($\delta$'s), and (b) MAD between SGE embedding and data, dented by $\mathcal{E}_S$,  for different neighborhood sizes and smoothing multiplier ($\mu$'s). (c) The difference of errors between these two methods ($\mathcal{E}_I-\mathcal{E}_S$) is computed in the variable space $\delta$ and $\mu_s$. \emph{The green cells denote that the performance of SGE is superior to that of Isomap.}}
	\label{fig:comp_embedding}
\end{figure}

The embedding error of SGE, denoted by $\mathcal{E}_S$, is computed as the Mean Absolute Deviation (MAD) between embedding and data \cite{petruccelli1999applied}. Since the distance matrices are symmetric and have zeros on the diagonal, MAD can then be computed using
\begin{equation} \label{eqn:mad}
\mathcal{E}_S=\frac{2}{n(n-1)}\sum^n_{i=1}\sum^n_{j=i+1}\big|(\bs{D}_M)_{ij}-(\bs{D}_S)_{ij}\big|.
\end{equation}
Similarly, for each pair of $\delta$ and $\mu_s$, we also compute MAD between the embedding of Isomap and the original data that we denote by $\mathcal{E}_I$. Fig.~\ref{fig:comp_embedding} illustrates MADs for Isomap ($\mathcal{E}_I$), MADs for SGE ($\mathcal{E}_S$), and their differences ($\mathcal{E}_I-\mathcal{E}_S$), versus $\delta$ and $\mu_s$. Fig.~\ref{fig:comp_embedding}(a) and \ref{fig:comp_embedding}(b) show that both methods display decreasing errors for increasing $\delta$'s (i.e., increasing neighbors). Moreover, SGE has a decreasing error as $\mu_s$ increases. Fig.~\ref{fig:comp_embedding}(c) also indicates that SGE performs better than Isomap for larger smoothing multipliers \emph{for all $\delta$'s}. This plot also shows that SGE performs worst when $\delta=2$ and $\mu_s=0$, and performs best when $\delta=2$ and $\mu_s=1$, as compared to isomap.

\begin{figure}[htp]
        	\centering
        	\includegraphics[width=3.2in]{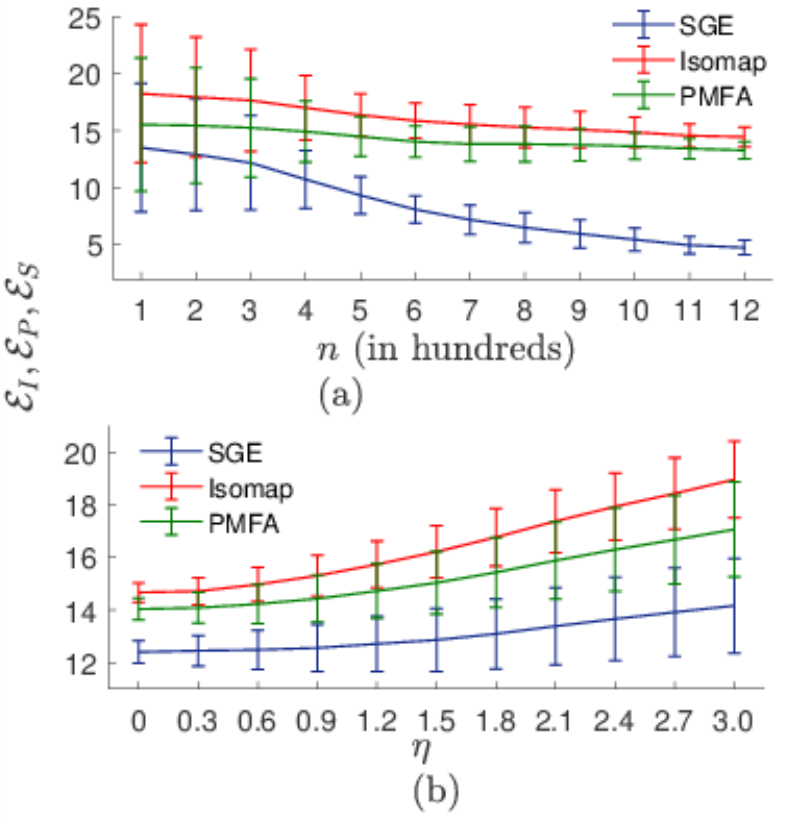}
        	\caption{Mean embedding error of Isomap (in red, denoted by $\mathcal{E}_S$), PMFA (in green, denoted by $\mathcal{E}_P$), and SGE (in blue, denoted by $\mathcal{E}_S$), versus, (a) sparcity and (b) noise. Error bars represent standard deviations of errors computed over realizations.  \emph{Note that, SGE has lower average error than that of Isomap and PMFA.}}
	\label{fig:analysis_n}
\end{figure}

Next, we analyze the influence of data sparsity on the manifold for embedding with SGE and compare this to PMFA and Isomap. For that task, we produce a sequence of spherical datasets with an increasing number of points.  We create the first dataset with 200 points using Eqn.~(\ref{eqn:sphere}) with $r=20+\mathcal{N}[0,2^2]$, then add another 100 points, generated using the same equation, into the first dataset to produce the second dataset. Similarly, we generate the last dataset of 1200 points. Then, we embed these datasets in two-dimensions using Isomap with $\delta=3$; PMFA with $n^1_c = n^2_c = 10$ and $s_r=0.9$; and SGE with $\delta=3$, $\mu_s=1$, $\nu=10\%$, and $h=100$. We compute the embedding errors $\mathcal{E}_I$, $\mathcal{E}_P$, and $\mathcal{E}_S$ using MAD for each dataset as explained before. Since a significantly high noise is used for the datasets, we create 16 such sequences of datasets and perform this analysis for16 realizations to allow us to compute averages. Fig.~\ref{fig:analysis_n}(a) shows the mean of embedding errors of 16 realizations and error bars for  SGE, Isomap, and PMFA. We observe that the error associated with SGE embedding is smaller than that of Isomap and PMFA for \emph{all the values of $n$}. This observation demonstrates the advantages of dealing with sparse data when comparing SGE to Isomap and PMFA.

Finally, we study the embedding errors of those three methods in terms of the size of the noise present in the data. For that task, we create a latticed semi-sphere of 600 points using Eqn.~(\ref{eqn:sphere}) with appropriately discretized $\gamma_1 \in[-\pi/2,\pi/2]$ and $\gamma_2 \in [0, \pi]$. Then, we impose increasing uniform noise levels into the parameter representing the radius as $r=20+\eta\text{U}[-1,1]$; $\eta=0, 0.3, 0.9, \dots, 3$ and produce a sequence of 11 datasets. We embed each dataset using Isomap with $\delta=3$; using PMFA with $n^1_c=n^2_c=10$ and $s_r=0.9$; and using SGE with $\delta=3$, $\nu=10\%$, $\mu_s=1$, and $h=100$. We create 25 such sequences of datasets and perform this analysis for 25 realizations to allow us to compute averages. Fig.~\ref{fig:analysis_n}(b) presents mean embedding errors and error bars for all three methods computed using Eqn.~(\ref{eqn:mad}). We observe that, while $\mathcal{E}_S$ slowly increases with increasing $\eta$, $\mathcal{E}_I$ and $\mathcal{E}_P$ increase quickly with increasing $\eta$.

\subsection{Embedding of face images}
\begin{figure}[hpt]
	\vspace{20pt}
	\begin{subfigure}{.5\textwidth}
        		\centering
        		\includegraphics[width=3.4in]{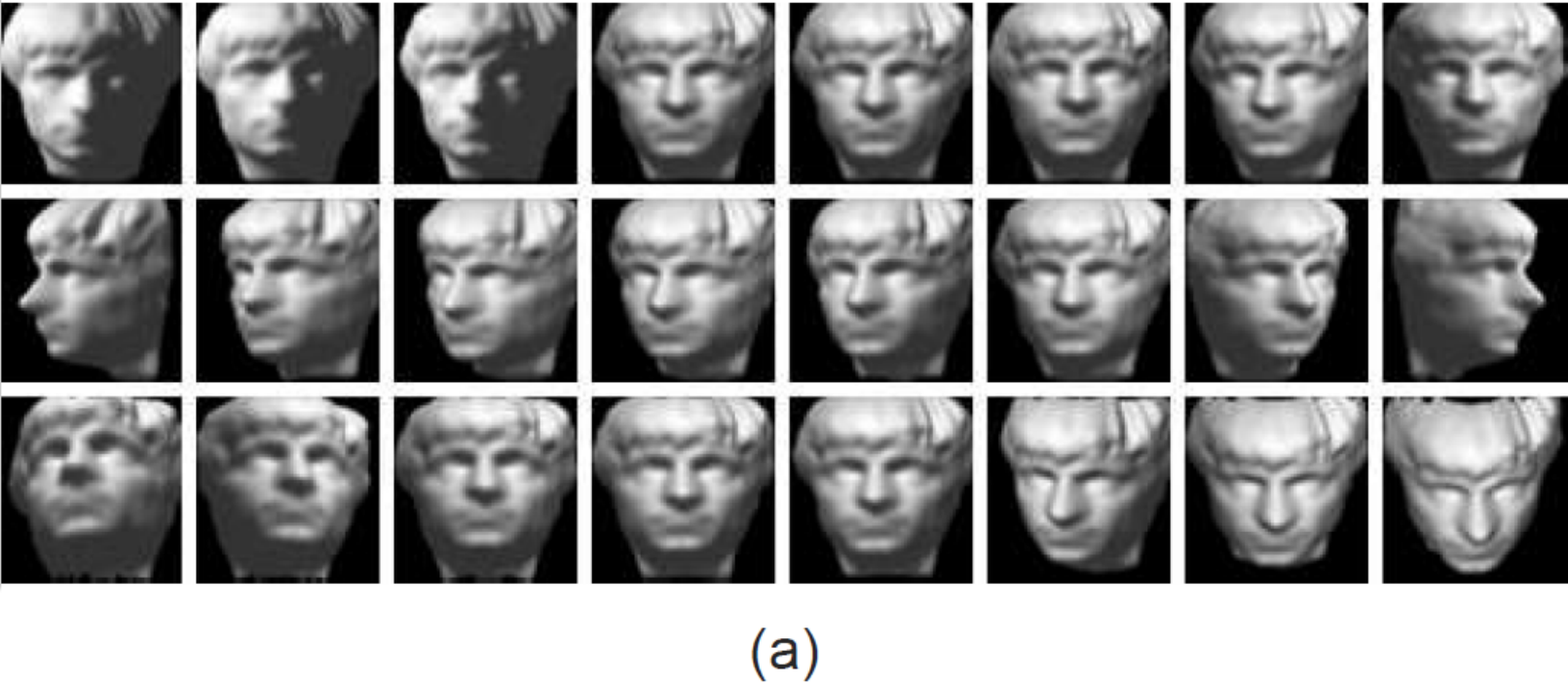}
    	\end{subfigure}
	~
	\begin{subfigure}{.5\textwidth}
        		\centering
        		\includegraphics[width=3.5in]{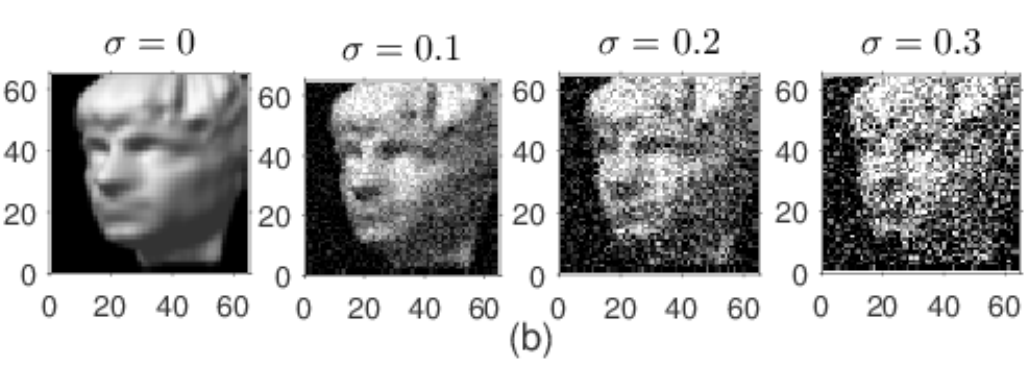}
    	\end{subfigure}
	~
	\begin{subfigure}{.5\textwidth}
        		\centering
        		\includegraphics[width=3.5in]{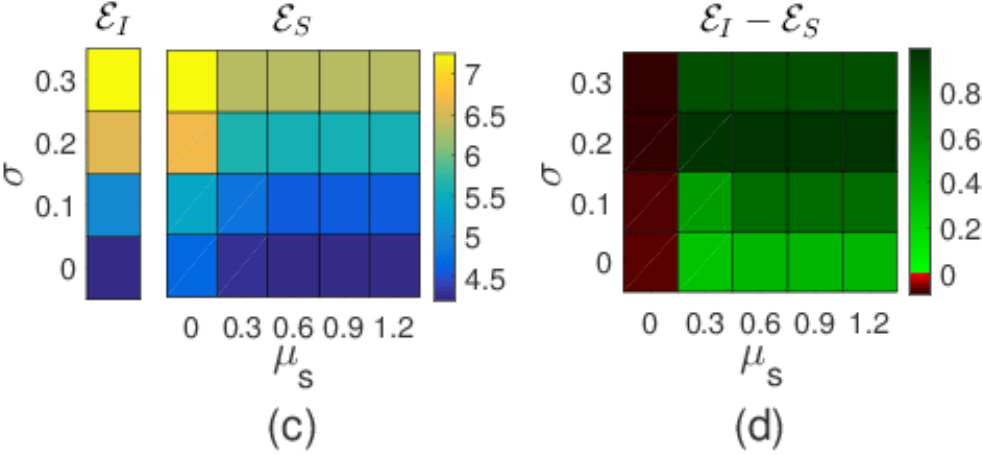}
	\end{subfigure}
	
	\caption{Embedding of face images ($64\times 64$ dimensional), distorted with different noise levels, using Isomap and SGE with different smoothing levels. (a) A sample of 16 face images \cite{deeplearning}; where the snapshots in the first, second, and third rows, represent left-right light variation, left-right pose variation, and up-down pose variation, respectively. (b) Face images are distorted by imposing three levels of Gaussian noise $\sigma=$ 0.1, 0.2, and 0.3. The datasets (four in total) are embedded using Isomap and then using SGE with smoothing multipliers $\mu_s=$ 0, 0.3 0.6, 0.9, 1.2. Then, (c) the embedding errors of Isomap ($\mathcal{E}_I$) and SGE ($\mathcal{E}_S$), and (c) their error difference are computed.  \emph{The green cells denote that the performance of SGE is superior to that of Isomap.}}\label{fig:face_images}
\end{figure}

In this section, we validate the SGE method using a real-world dataset of face images available in \cite{deeplearning}. This dataset consists 698 images each of $64\times 64$ dimension with a varying pose and direction of lighting, as shown by a sample of 16 snapshots in Fig.~\ref{fig:face_images}(a). We randomly choose 400 images as our baseline dataset and generate three other datasets of 400 images from the baseline dataset by imposing Gaussian noise with standard deviations ($\sigma's$) 0.1, 0.2, and 0.3 [Fig.~\ref{fig:face_images}(b)]. We set $\delta=4$, $\nu=10\%$, $h=100$ in SGE and run this algorithm over each dataset (4 in total) 5 times for $\mu_s= $0, 0.3, 0.6, 0.9, 1.2. Then, we embed these four datasets in two dimensions using Isomap and LSE with $\delta=4$.

We use the ability to preserve distances between the original and the embedding data to analyze the performance of the method \cite{shaw2009structure}. In particular, we consider the distances in the original (noise free) imagery as the ``true'' distances and judge the algorithm's ability to recover those distances after the imagery has been corrupted by noise.For both the data and the embedding, we first search $\delta$ nearest neighbors for each point and then produce a weighted graph by treating points in the dataset as nodes and connecting each two neighbors by an edge having the length equal to their Euclidean distance. The weighted graph constructed through the nearest neighbor search is a simple graph\footnote{A simple graph is an undirected graph that does not contain loops (edges connected at both ends to the same vertex ) and multiple edges (more than one edge between any two different vertices) \cite{balakrishnan2012textbook}.} that does not contain self-loops or multiple edges. We compute the $ij$-th entry of the adjacency distance matrix $A$ for the data as
\begin{equation}\label{eqn:adj1}
A_{ij}= \begin{cases}
      	d(i,j) 	& : \text{if} \ \exists \ \text{an edge} \  ij \ \text{in the graph} \\
           	&  \ \ \text{of the original data,}\\
      	0 	& : \text{otherwise,}
   \end{cases}
\end{equation}
and the $ij$-th entry of the adjacency distance matrix $\tilde{A}$ for the embedding data as
\begin{equation}\label{eqn:adj2}
\tilde{A}_{ij}= \begin{cases}
      	d(i,j) 	& : \text{if} \ \exists \ \text{an edge} \  ij \ \text{in the graph} \\
		&  \ \ \text{of the embedding data,}\\
      	0 	& : \text{otherwise.}
   \end{cases}
 \end{equation}
Here, $d(i,j)$ is the Euclidean distance between nodes $i$ and $j$. In this paper, we impose Gaussian noise into real-world datasets such as face images and images of handwritten digits (Sec.~\ref{sec:handDigit}). \emph{Thus, we think of our original data as the uncorrupted data before we impose the noise.}

For $n$ points in the dataset, the error associated with the neighbors' distance is computed as the average of absolute differences between entries of the adjacency distance matrices,
\begin{equation}\label{eqn:adj_error}
\mathcal{E}=\frac{1}{n(n-1)}\sum_{i,j=1}^{n}\big\vert A_{ij}-\tilde{A}_{ij}\big\vert,
\end{equation}
where $\delta$ is the neighbor parameter \cite{shaw2009structure}.

Fig.~\ref{fig:face_images}(c) illustrates the embedding errors of Isomap, denoted by $\mathcal{E}_I$, and  embedding errors of SGE, denoted by $\mathcal{E}_S$, for $\sigma= $ 0, 0.1, 0.2, 0.3 and $\mu_s= $ 0, 0.3, 0.6, 0.9, 1.2.  We observe that the error increases in both methods when the noise in the data increases. However, the error of embedding noisy data can be reduced significantly by choosing an appropriate smoothing multiplier in SGE as shown here. Fig.~\ref{fig:face_images}(d) showing the difference of errors ($\mathcal{E}_I$-$\mathcal{E}_S$) demonstrates that SGE performs better in terms of error than Isomap for all the noise levels with $\mu_s \ge 0.3$.

\subsection{Embedding of handwritten digits}\label{sec:handDigit}
Next, we embed handwritten digits available from the Mixed National Institute of Standards and Technology (MNIST) database \cite{lecun2016the} using SGE and study the performance of the method. This dataset contains 60,000 images of handwritten digits from 0 to 9 each of $28 \times 28$ dimensions. We sample two arbitrary datasets for our study, each with 400 images, such that one dataset has only the digit 2 and the other dataset has the digits 2, 4, 6, and 8.

We run Isomap over the dataset having the digit 2 with $\delta=4$. We run SGE over this dataset two times: first with $\delta=4$, $\mu_s=0$, $\nu=10\%$, and $h=100$; and second with $\delta=4$, $\mu_s=0.6$, $\nu=10\%$, and $h=100$. Thus, aforesaid procedure yields three two-dimensional embeddings. We formulate the adjacency distance matrices for the data and embedding using Eqns.~(\ref{eqn:adj1}) and (\ref{eqn:adj2}), respectively, and compute the error of embedding using Eqn.~(\ref{eqn:adj_error}). Then, we distort this dataset with a Gaussian noise having $\sigma=0.2$ and run Isomap with $\delta=4$. We run the noisy dataset two times in SGE: first with parameters $\delta=4$, $\mu_s=0$, $\nu=10\%$, and $h=100$; and second with $\delta=4$, $\mu_s=0.6$, $\nu=10\%$, and $h=100$. The embedding errors for Isomap, SGE with $\mu_s=0$, and SGE with $\mu_s=0.6$, are given in Table~\ref{tab:mnist}(a). We see in this table that, regardless of the noise present in the data, the error associated with SGE without smoothing is greater than that of Isomap, while that of SGE with smoothing is smaller than that of Isomap. Moreover, the error of embedding is increased when moving from the noisy dataset to the noise free dataset by 0.87 for Isomap, while that is only increased by 0.24 for SGE with $\mu_s=0.6$.  This is due to the fact that setting the smoothing multiplier $\mu_s$ to $0.6$ allows SGE to recover the manifold corrupted by noisy measurements.

Next, we embed a sample of 400 digits, consisting of 2's, 4's, 6's, and 8's, into two dimensions using Isomap and SGE. We run Isomap over this dataset with $\delta=4$. Then, run SGE two times: first with $\delta=4$, $\nu=10\%$, $\mu_s=0$, and $h=100$; and second with $\delta=4$, $\nu=10\%$, $\mu_s=0.9$, and $h=100$. Thereafter, we distort the dataset with a Gaussian noise having $\sigma=0.3$ and then run Isomap with $\delta=4$ followed by running SGE with the same two parameter sets that we used before. Then, we compute the Isomap and SGE errors associated with embedding of noise free and noisy datasets using Eqn.~(\ref{eqn:adj_error}) that we present in  Table~\ref{tab:mnist}(b). Similarly to the embedding of the digit 2, regardless of the error in the data, here we also note that the embedding error of SGE \emph{with no smoothing} is greater than that of Isomap, while the embedding error of SGE \emph{with smoothing} is smaller than that of Isomap. Moreover, moving from embedding of noise free data to embedding of noisy data, while the error associated with Isomap is increased by 0.88, that of SGE with $\mu_s=0.9$ is increased only by 0.21.

\begin{figure}[htp]
        	\centering
        	\includegraphics[width=3.5in]{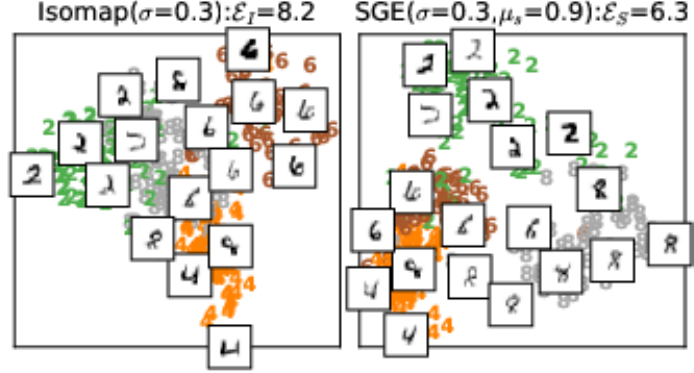}
        	\caption{Isomap and SGE embeddings of handwritten numbers 2, 4, 6, and 8. Different digits are shown in different colors (2's in green, 4's in orange, 6's in brown, and 8's in gray) in the two-dimensional embedding space and the embedded snapshots illustrate the appearance of arbitrarily chosen handwritten digits. The left panel shows the Isomap embedding of the noisy dataset ($\sigma=0.3$) and the right panel shows the SGE embedding of the same dataset with $\mu_s=0.9$. The embedding error of each case is indicated in the title of the corresponding panel. Note that, qualitatively speaking, the SGE embedding appears to have better clustering of similar digits than that of Isomap.}
	\label{fig:handwring_2468}
\end{figure}

Finally, we compare the classification ability of both methods in the presence of high noise. In this example, we define classification as spatially clustering of similar digits. To visualize the classification ability, we construct two-dimensional Isomap and SGE embeddings of the noisy dataset ($\sigma=0.3$) of digits 2, 4, 6, and 8, that we present in Fig.~\ref{fig:handwring_2468}. Therein, we observe that, while Isomap's embedding is unable to maintain clear boundaries between clusters of the same digit, SGE could at least separate numbers 2 and 8 from the rest of the digits. Thus, we can conclude that while Isomap is unable to achieve a clear classification of digits, SGE with $\mu_s=0.9$ achieves qualitatively better classification, even under the high noise present in the data.

\begin{table}[htp]
\begin{center}
\begin{tabular}{|c|c|c|c|c|}
\hline
\multirow{2}{*}{(a)} & \multirow{2}{*}{Noise} & \multirow{2}{*}{Isomap} & \multicolumn{2}{|c|}{SGE}\\
\cline{4-5}
& &  & $\mu_s=0$ & $\mu_s=0.6$\\
\hline
\multirow{2}{*}{Digit ``2"} & $\sigma=0$ &  7.02 & 7.13 & 5.86\\
\cline{2-5}
& $\sigma=0.2$ & 7.89 & 8.19 & 6.10\\
\hline \hline
\multirow{2}{*}{(b)} & \multirow{2}{*}{Noise} & \multirow{2}{*}{Isomap} & \multicolumn{2}{|c|}{SGE}\\
\cline{4-5}
& & & $\mu_s=0$ & $\mu_s=0.9$\\
\hline
Digits ``2", ``4", & $\sigma=0$ & 7.38 & 7.43 & 6.09\\
\cline{2-5}
``6", and ``8" & $\sigma=0.3$ & 8.20 & 8.36 & 6.30\\
\hline
\end{tabular}

\end{center}
\caption {Errors of Isomap and SGE embeddings of, (a) a sample of 400 handwritten 2's; and (b) a sample of 400 handwritten digits having number 2's, 4's, 6's, and 8's. The first row of (a) shows the error when the  dataset of digit 2 is embedded using both Isomap, and SGE with two smoothing coefficients $\mu_s=0$ and $\mu_s=0.6$. Then, the dataset is imposed with a Gaussian noise of $\sigma=0.2$ and embedded using both Isomap, and SGE with $\mu_s=0$ and $\mu_s=0.6$ that you see in the second row of (a). The first row of (b) represents the errors of Isomap embedding, and SGE embeddings with $\mu_s=0$ and $\mu_s=0.9$, of the noise free version of the sample of digits having the numbers 2, 4, 6, and 8. The second row of (b) represents the errors of Isomap embedding, and SGE embeddings with $\mu_s=0$ and $\mu_s=0.9$, of the noisy version of the dataset created by imposing a Gaussian noise with $\sigma=0.3$.} \label{tab:mnist}
\end{table}

\section{Conclusion}\label{sec:conclusion}
Nonlinear dimensionality reduction methods can recover unfaithful embeddings due to sparsity and presence of  high noise in the data.  In order to obtain a faithful embedding for sparse and noisy data, some smoothing procedure should be performed in the embedding. With this idea in mind, herein we introduced a novel nonlinear dimensionality reduction framework using smooth geodesics that emphasizes the underlying smoothness of the manifold. Our method begins by first searching for nearest neighbors for each point using a $\delta$-nearest neighbor search \cite{friedman1977algorithm}. Then, we create a weighted graph by representing all of the data points as nodes and joining neighboring nodes with edges having their Euclidean distances as weights. For each pair of nodes in the graph, we create a geodesic \cite{tenenbaum2000global}, that is defined as the shortest path between the given nodes, generated using Floyd's algorithm \cite{cormen2009introduction}. We replaced Dijkstra's algorithm in classic Isomap with Floyd's algorithm since Floyd's algorithm is more computationally efficient than Dijkstra's algorithm in our case. We fit each such geodesic with a smoothing spline (called a smooth geodesic) that is controlled by two parameters: smoothing multiplier ($\mu_s$) and spline threshold ($\nu$) \cite{reinsch1967smoothing, de1972calculating}. The length of these splines are treated as manifold distances between corresponding points. 

We use a classic MDS method to find the dimension of the distance matrix of smooth geodesics and perform the embedding. The MDS method converts the squared distance matrix to a Gram matrix, employs EVD to compute eigenvalues and eigenvectors, and finally uses Eqn.~\ref{eqn:latent_var} to produce the embedding \cite{lee2007nonlinearb}. In theory, a Gram matrix of a squared distance matrix should be SPD \cite{lee2007nonlinearb}. However, due to geodesic approximation of the true manifold distance and other numerical approximations, this Gram matrix might deviate slightly  from being SPD. Since this slightly corrupted Gram matrix produces small negative eigenvalues and those then violate Eqn.~\ref{eqn:latent_var}, we replace EVD in MDS by SVD. Note that, the EVD and SVD of an exact Gram matrix are the same.

In SGE, the order of the spline fit is set to either three, two, or one, depending on the spline threshold. Since a sufficient smoothness and a low fitting error can be obtained by cubic smoothing splines, we first rely on a spline fit of  order three. However, we observed that the smoothing spline routing in \cite{de1972calculating} fits very long cubic splines for some specific smoothing multipliers. Thus, if the length of a cubic smoothing spline doesn't satisfy the threshold, we reduce the order of the spline to a lower level. Choosing the order of the spline can also be considered as a trial and error process that we implement here by utilizing a threshold. In future, we will consider replacing this threshold by a trial and error process to obtain a faithful embedding. Hereby, we also be able to consider higher orders for the spline in Eqn.\eqref{eqn:spline} than the current highest order of three.

The smoothing spline approach in SGE approximates the true manifold more accurately in many cases than the geodesic approach in Isomap (Fig.~\ref{fig:spl_approximation}). Specifically, the use of smoothing splines can closely approximates the true manifold distance even when the data is sparse, in contrast to Isoamp which creates polygonal paths that then add extra length to the true manifold distance. In the limit of infinite number of sample points, the Isomap geodesics converge to the true manifold distance\cite{bernstein2000graph}. However, due to the smoothing spline approach in SGE, we have observed that the embedding error of SGE is significantly lower than that of Isomap in many practical problems. This was evidenced by the semi-sphere example in Sec.~\ref{sec:semi-sphere}. Although, both methods do not converge to the true manifold when the data is corrupted by noise, SGE emphasizes the smoothness of the manifold while Isomap is severely impacted as the errors in the lengths of the Isomap polygonal paths is intensified.

In the future, we plan to modify the SGE method such that it converges even when the data is corrupted by noise. For that, first, we will need to estimate, or be provided with, the curvature of the manifold described by noisy data. Then, we plan to replace the nearest neighbor search with a range search \cite{agarwal1999geometric} that finds all the points within a given distance based on the curvature. We will run the range search over the points those are close to the manifold (we can find these points as we know the curvature), and then create a graph by treating data points as nodes and connecting neighbors by edges. We expect that the range search will ensure that the graph doesn't have long edges arising from highly noisy points in data. As the range search is ran over the data points close to the manifold, we expect to be able to create shortest paths those are close to the manifold. Finally, we will follow all the other steps 3--6 as in Algorithm \ref{alg:algorithm} of SGE and compute the embedding.

We first demonstrated the effectiveness of our NDR method on a synthetic dataset representing a semi-sphere. We observed that the smoothing approach provides a better embedding performance than the embedding achieved by standard Isomap or PMFA when embedding a noisy dataset. We also observed that the errors in Isomap and SGE decrease as the neighborhood size increases [Fig.~\ref{fig:comp_embedding}(a) and (b)].  However, when the neighborhood size is small, say $\delta=2$, SGE has clear performance advantages over Isomap for noisy data when a sufficient smoothness is employed [Fig.~\ref{fig:comp_embedding}(c)]. The spherical dataset also demonstrated that SGE is more robust to sparse sampling than Isomap and PMFA [Fig.~\ref{fig:analysis_n}(a)]. Moreover, while increasing noise in the data always appears to reduce the performance of the embedding, irrespective to the method that is used, we see that Isomap and PMFA are highly effected by increasing noise while SGE, with a judicious choice of smoothing multiplier, is more robust [Fig.~\ref{fig:analysis_n}(b)].

We also studied two standard benchmark datasets, face images \cite{deeplearning} and handwritten digit images \cite{lecun2016the}, and found that SGE provided similar superior performance on noisy versions of those datasets. In particular, for the digit classification task, we observed that SGE provides qualitatively superior classification performance (that is, clustering similar digits into one group) in the presence of noise. As future work, we will quantify the classification performance of the low-dimensional nonlinear embedding using a variety of standard supervised machine learning techniques.

As a potential application of SGE, we can modify some semi-supervised learning methods such as \cite{Wang2009a} and metric learnings such as \cite{wang2015psf} by replacing their LDR routing, applied in unlabeled portions of their data, by SGE. By doing this, we believe that the results of those methods can be improved when they are applied to nonlinear data due to the nonlinear and smoothing features of SGE. Specifically, the similarity evaluation framework in \cite{wang2015psf} inputs both supervised and unsupervised data. The data here is a collection of patients' claim records, lab test records, and pharmacy records. A distance measure between each pair of patients is generated as the sum of the distances in the supervised data and the unsupervised data. The distances in unsupervised data are computed on the embedding using LDR methods, such as Local Linear Embedding (LLE) \cite{roweis2000nonlinear} and Laplacian eigenmaps \cite{belkin2003laplacian}, might not emphasis nonlinear features of the data. Thus, replacing these LDR method by SGE can produce accurate results in large array of datasets. On the other hand, the graphed based classification routing presented in \cite{Wang2009a} leverage semi-supervised data (data with labels and without labels) and creates a graph structure of the data. Then, it estimates the edge weights using a modified LLE method followed  by performing a classification which is done based edge weights. However, using a LDR method such as LLE for nonlinear data is not very efficient as it does not leverage the nonlinear features of the data. We believe that replacing the LLE routing in this method by SGE can produce improved results in many such datasets. 

The NDR method that we introduced here ensures better performance and preserves the topology of the manifold by emphasizing the smoothness of the manifold when embedding sparse and noisy data. This method is an extension of famous NDR method Isomap where we replaced the geodesics with smoothing splines.  In the future, we plan to examine such techniques in more generally.  For example, one can imagine generalizing Isomap to the case where both geodesics and smoothing splines are not a good approximation of long manifold distances.  In such a case, one can attempt to treat the long manifold distances as unknown, and employ matrix completion techniques like \cite{lin2010augmented, paffenroth2013space} on distance matrices where some entries are not observed.


%


\ifCLASSOPTIONcompsoc
  \section*{Acknowledgments}
\else
  \section*{Acknowledgment}
\fi
The author's would like to thank Chen Zou for the support given in coding and
also would like to thank the NSF XSEDE Jetstream
\cite{stewart2015jetstream,towns2014xsede} under allocation TG-DMS160019 for
support of this work.
\ifCLASSOPTIONcaptionsoff
  \newpage
\fi



\bibliographystyle{IEEEtran}




%

\begin{IEEEbiography}[{\includegraphics[width=1in,height=1.25in,
clip,keepaspectratio]{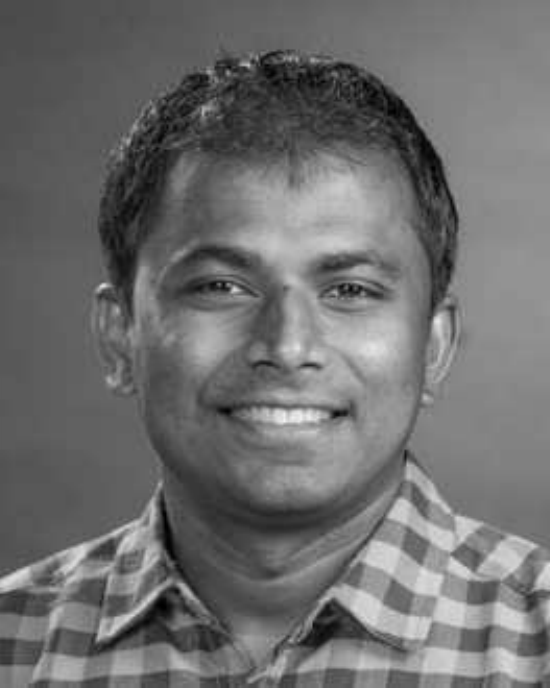}}]{Kelum Gajamannage}
Dr. Kelum Gajamannage was awarded his BS (with honor) in Mathematics and MS in Applied Statistics from University of Peradeniya, Sri Lanka. He worked three years as a lecturer in the Department of Science and Technology at Uva Wellassa University, Sri Lanka. Dr. Gajamannage graduated with a PhD in Mathematics at Clarkson University, USA. He is currently a PostDoctoral Scholar in the department of Mathematical Sciences, Worcester Polytechnic Institute, USA.
\end{IEEEbiography}

\begin{IEEEbiography}[{\includegraphics[width=1in,height=1.25in,
clip,keepaspectratio]{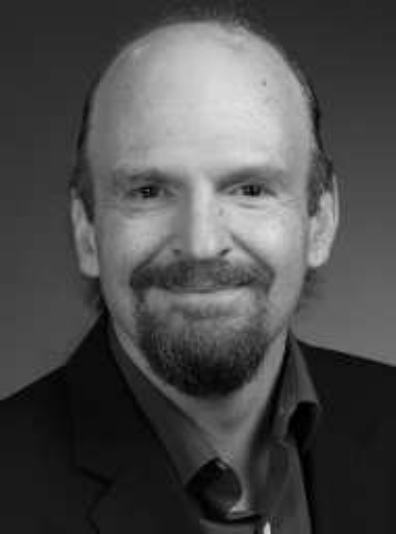}}] {Randy Paffenroth} Dr.
Paffenroth graduated from Boston University with degrees in both mathematics and
computer science and he was awarded his Ph.D. in Applied Mathematics from the
University of Maryland in June of 1999.  After attaining his Ph.D., Dr.
Paffenroth spent seven years as a Staff Scientist in Applied and Computational
Mathematics at the California Institute of Technology. In 2006, he joined
Numerica Corporation where he held the position of Computational Scientist and
Program Director.  Dr. Paffenroth is currently an Associate Professor of
Mathematical Sciences and Associate Professor of Computer Science at Worcester
Polytechnic Institute where his focus is on the WPI Data Science Program.  His
current technical interests include machine learning, signal processing, large
scale data analytics, compressed sensing, and the interaction between
mathematics, computer science and software engineering, with a focus on
applications in cyber-defense. \end{IEEEbiography}


\begin{IEEEbiography}[{\includegraphics[width=1in,height=1.25in,
clip,keepaspectratio]{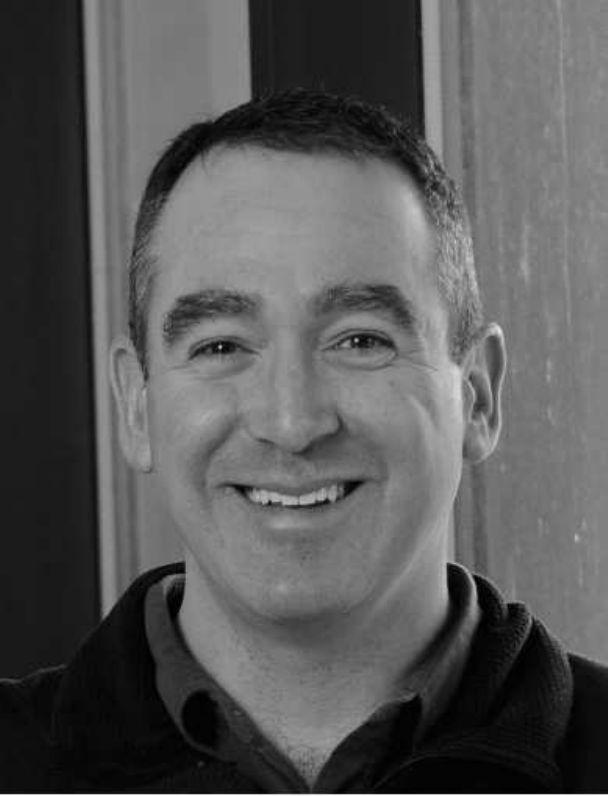}}]{Erik M. Bollt} Dr. Erik Bollt was awarded his Ph.D. in Mathematics from the
University of Colorado in Boulder. Currently, he is a Full Professor at Clarkson University and endowed as the W. Jon Harrington Professor of Mathematics.  Professor Bollt specializes in dynamical systems, chaos theory and turbulence, including as informed by data and signal processing, and remote sensing of the world’s oceans, as well as inverse problems in data processing and information theoretic questions for information flow and systems inference.  Prof. Bollt has recently published a book on these topics as applied to systems such as the Gulf of Mexico oil spill, [Applied and Computational Measurable Dynamics, Book Publisher: Society for Industrial and Applied Mathematics, (2013)].
\end{IEEEbiography}




\end{document}